\documentclass[10pt,twocolumn,letterpaper]{article}

\usepackage{iccv}
\usepackage{times}
\usepackage{epsfig}
\usepackage{graphicx}
\usepackage{amsmath}
\usepackage{amssymb}
\usepackage{multirow}
\usepackage{threeparttable}
\usepackage{bbding}
\usepackage{bm}

\usepackage[pagebackref=true,breaklinks=true,letterpaper=true,colorlinks,bookmarks=false]{hyperref}
\usepackage{array}
\newcommand{\PreserveBackslash}[1]{\let\temp=\\#1\let\\=\temp}
\newcolumntype{C}[1]{>{\PreserveBackslash\centering}p{#1}}
\newcolumntype{R}[1]{>{\PreserveBackslash\raggedleft}p{#1}}
\newcolumntype{L}[1]{>{\PreserveBackslash\raggedright}p{#1}}

\iccvfinalcopy

\def\iccvPaperID{random}

\ificcvfinal\fi

\begin{document}

\title{Reinforcing Local Feature Representation for Weakly-Supervised Dense Crowd Counting}

\author{Xiaoshuang Chen\\
{\tt\small chenxiaoshuang@sjtu.edu.cn}
\and
Hongtao Lu\\
{\tt\small htlu@sjtu.edu.cn}
}

\maketitle

\ificcvfinal\thispagestyle{empty}\fi

\begin{abstract}
Fully-supervised crowd counting is a laborious task due to the large amounts of annotations.
Few works focus on weekly-supervised crowd counting, where only the global crowd numbers are available for training.
The main challenge of weekly-supervised crowd counting is the lack of local supervision information.
To address this problem, we propose a self-adaptive feature similarity learning (SFSL) network and a global-local consistency (GLC) loss to reinforce local feature representation.
We introduce a feature vector which represents the unbiased feature estimation of persons.
The network updates the feature vector self-adaptively and utilizes the feature similarity for the regression of crowd numbers.
Besides, the proposed GLC loss leverages the consistency between the network estimations from global and local areas.
The experimental results demonstrate that our proposed method based on different backbones narrows the gap between weakly-supervised and fully-supervised dense crowd counting.
\end{abstract}

\section{Introduction}

Crowd Counting is an essential topic which has been widely applied in many tasks, such as video surveillance, urban traffic management, and safety monitoring.
Density estimation based methods~\cite{zhang2016single,li2018csrnet,cao2018scale,jiang2019crowd,bai2020adaptive,jiang2020attention,song2021choose} and point annotation based methods~\cite{ma2019bayesian,wang2020distribution,ijcai2021-119,song2021rethinking} have achieved remarkable performance.

However, existing fully-supervised methods for crowd counting require large amounts of annotations.
For example, the UCF-QNRF~\cite{idrees2018composition} datasets contains annotations of 1.25 million people, where there are 815 labels per image on average, and the NWPU-Crowd~\cite{wang2020nwpu} datasets contains annotations of 2.13 million people, where there are 418 labels per image on average.
The location of each objects in an image or the density map of an image serves as the supervision target when training the counting network.
The annotation work is laborious and time-consuming, and the evaluation metric of crowd counting does not take the location information into account.

There is a significant reduction in the annotation amount in semi-supervised methods~\cite{olmschenk2019dense,liu2020semi,meng2021spatial,xu2021crowd}, where only limited location annotations are used in the training process, e.g., annotations of limited images~\cite{olmschenk2019dense,liu2020semi,meng2021spatial} or partial annotations in each image~\cite{xu2021crowd}.
However, the location annotations are still required to supervise the network training.
Actually the crowd numbers in images can be obtained by other approaches, such as crowd sensing technology~\cite{guo2015mobile}.
Sheng et al.~\cite{sheng2014leveraging} leverages GPS-less energy-efficient sensing scheduling for mobile crowd sensing technology.
Besides, the crowd number of the same scene keeps constant with different viewpoints in many cases, so repeated annotating is unnecessary.
Yang et al.~\cite{yang2020weakly} is the first to propose a weakly-supervised framework to regress the crowd number directly and exploits the relationship among different images by a sorting network.
But it is incapable of addressing large count variance in the datasets.
The most challenging problem of weakly-supervised crowd counting is the lack of local feature representation.
Briefly, the extracted features should be more discriminative in the local areas.

In this paper, we propose a self-adaptive feature similarity learning (SFSL) network and a global-local consistency (GLC) loss to reinforce local feature representation.
In the weakly-supervised crowd counting task, there is insufficient local constraint in the supervision information.
Our contributions based on this prerequisite are two-fold.

Firstly, considering the prediction of crowd counting is similar with the pixel-wise soft binary classification task, we introduce an unbiased feature estimation for the positive samples.
This unbiased feature estimation can be regarded as the cluster center of the positive class, i.e., the class of persons.
However, the local annotations are unavailable in the weakly-supervised manner, so it is difficult to classify the positive and the negative samples.
In our training process, this feature vector is updated self-adaptively by gradient descent.
We calculate the feature similarity between the unbiased feature estimation and the feature vectors extracted from the backbone network at each position and utilize the similarity for the regression of crowd numbers.
The similarity map is considered as the soft binary classification probability to construct the density map, and is also used as a part of the input vector of the linear regression network to regress the final crowd number as well.

Secondly, only regressing the crowd number at the global level ignores the estimations at the local level and may lead to suboptimal predictions.
Based on the fact that the crowd number in the global area should be equivalent to the summation of those in the local areas, we propose a global-local consistency (GLC) loss to enforce the consistency between the predictions at the global and the local level.
We divide the original image into multiple subimages, each of which contains a local area.
The regression of local areas is trained in a self-supervised manner, i.e., supervised by the global predictions of crowd numbers.
Leveraging the semantic information in local regions, the network can extract more discriminative features from limited data.

We carry out experiments on four crowd counting datasets and compare the results with fully-supervised methods which need the location labels.
Besides, taking the labeling error brought by inaccurate annotation approaches into account, we test the robustness of our method under the circumstance of labeling deviations of crowd numbers.
Furthermore, our method has general applicability, improving the performance of baselines based on different backbones, including CNN and ViT~\cite{dosovitskiy2020image}.

The main contributions of this paper are summarized as follows:

\begin{itemize}
    \item We propose a self-adaptive local feature similarity learning network.
    We introduce an unbiased feature estimation and utilize the pixel-wise feature similarity for the regression of crowd numbers.
    \item We propose a global-local consistency (GLC) loss to leverage the consistency between the network estimations from global and local areas by supervising the local regression with the global estimations.
    \item Sufficient experiments demonstrate that our proposed method narrows the gap between weakly-supervised and fully-supervised dense crowd counting on datasets with large count variance.
    Besides, our method is robust to the labeling deviations of crowd numbers.
\end{itemize}

\section{Related Works}

\subsection{Fully-Supervised Methods}

\subsubsection{Density Estimation Based Methods}

Fully-supervised density estimation based methods, which generate density maps, are the mainstream methods for crowd counting.
The ground truth density maps are constructed by employing Gaussian kernels to the location maps.
For example, CSRNet~\cite{li2018csrnet}, which employs dilated convolution to enlarge the receptive fields, is a common baseline method for density estimation.

There are lots of methods aiming at the scale variation problem in crowd counting.
MCNN~\cite{zhang2016single} uses three branches with multi-size convolution kernels to address the scale variation problem.
SANet~\cite{cao2018scale} extracts features with different receptive fields using scale aggregation modules.
TEDNet~\cite{jiang2019crowd} is a trellis encoder-decoder network that contains multiple decoding paths to aggregate features.
PGCNet~\cite{yan2019perspective} introduces an effective perspective estimation branch to overcome the scale variation due to the perspective effect.
ADSCNet~\cite{bai2020adaptive} proposes an adaptive dilated convolution and a self-correction supervision framework to address the large scale variation and labeling deviations.
There are other methods using auxiliary tasks such as the segmentation task to assist the counting task.
Zhao et al.~\cite{zhao2019leveraging} utilizes the geometric, semantic and numeric attributes to formulate auxiliary tasks.
Shi et al.~\cite{shi2019counting} proposes focus from segmentation, global density and improved density maps.
Besides, there are also some methods~\cite{liu2019adcrowdnet,guo2019dadnet,zhang2019attentional,zhang2019relational,jiang2020attention} introduce attention mechanism for more accurate counting.

\subsubsection{Point Annotation Based Methods}

Instead of generating density maps as the supervision targets, recently there are some point annotation based methods, which use the annotated points as the supervision targets.
Each annotated point corresponds to the location of one object.
Some of these methods outperform the density estimation based methods.

Bayesian Loss~\cite{ma2019bayesian} is a loss function which constructs a density contribution probability model from the point annotations.
DM-Count~\cite{wang2020distribution} uses Optimal Transport to measure the similarity between the predicted density map and the ground truth.
BM-Count~\cite{ijcai2021-119} proposes a bipartite matching based method and a new ranking distribution loss using point supervision.
P2PNet~\cite{song2021rethinking} proposes a purely point-based framework for crowd counting and localization, which directly predicts point proposals.

Fully-supervised methods using whether density maps or point annotation as the supervision targets require a large amount of annotations.
Most crowd counting datasets contain millions of objects, and the location of each objects is required for training in fully-supervised methods.
The annotation work is laborious and time-consuming, and the location information is not taken into account in the evaluation metric of crowd counting.
Aiming to address this problem, semi-supervised and weakly-supervised methods have attracted increasing attention recently.

\subsection{Semi-Supervised Methods}

In semi-supervised methods, only limited location annotations are available in the training process.
These methods reduce the annotation amount in crowd counting to some extent.

Some methods use annotations of limited images in the training sets and other methods use partial annotations in each image.
DG-GAN~\cite{olmschenk2019dense} presents a dual-goal GAN architecture using limited labeled data.
IRAST~\cite{liu2020semi} introduces surrogate tasks and develops a self-training method with much fewer density map annotations.
SUA~\cite{meng2021spatial} proposes a spatial uncertainty-aware semi-supervised approach, where only some images in the training sets are labeled.
Xu et al.~\cite{xu2021crowd} proposes Partial Annotation Learning only using partial annotations in each image as training data.

However, the amount of annotations is still large, because the location information is still required for training.
Actually, only the total crowd number of an image is taken into account in the evaluation metric of crowd counting.
Thus without all the location information, the annotation amount can be further reduced.

\subsection{Weakly-Supervised Methods}

Considering only the total crowd number is taken into account when evaluating the counting accuracy, weakly-supervised methods predict the total crowd number of each image directly without any location annotation.
Yang et al.~\cite{yang2020weakly} is the first to propose a weakly-supervised framework for crowd counting. This method regresses the crowd number directly and exploits the relationship among images by a sorting network.
MATT~\cite{lei2021towards} develops a Multiple Auxiliary Tasks Training strategy for weakly-supervised crowd counting.
TransCrowd~\cite{liang2021transcrowd} adopts Transformer~\cite{vaswani2017attention} architecture as the counting network to boost weakly-supervised crowd counting.

However, without location annotations, the local feature representation is weakened for the lack of local supervision information.
To address the problem, we propose a self-adaptive feature similarity learning (SFSL) network and a global-local consistency (GLC) loss to reinforce local feature representation and thus boost the performance of weakly-supervised crowd counting with only the total crowd numbers as training labels.

\section{Method}

\begin{figure*}[tp]
    \centering
    \includegraphics[width=1.0\linewidth]{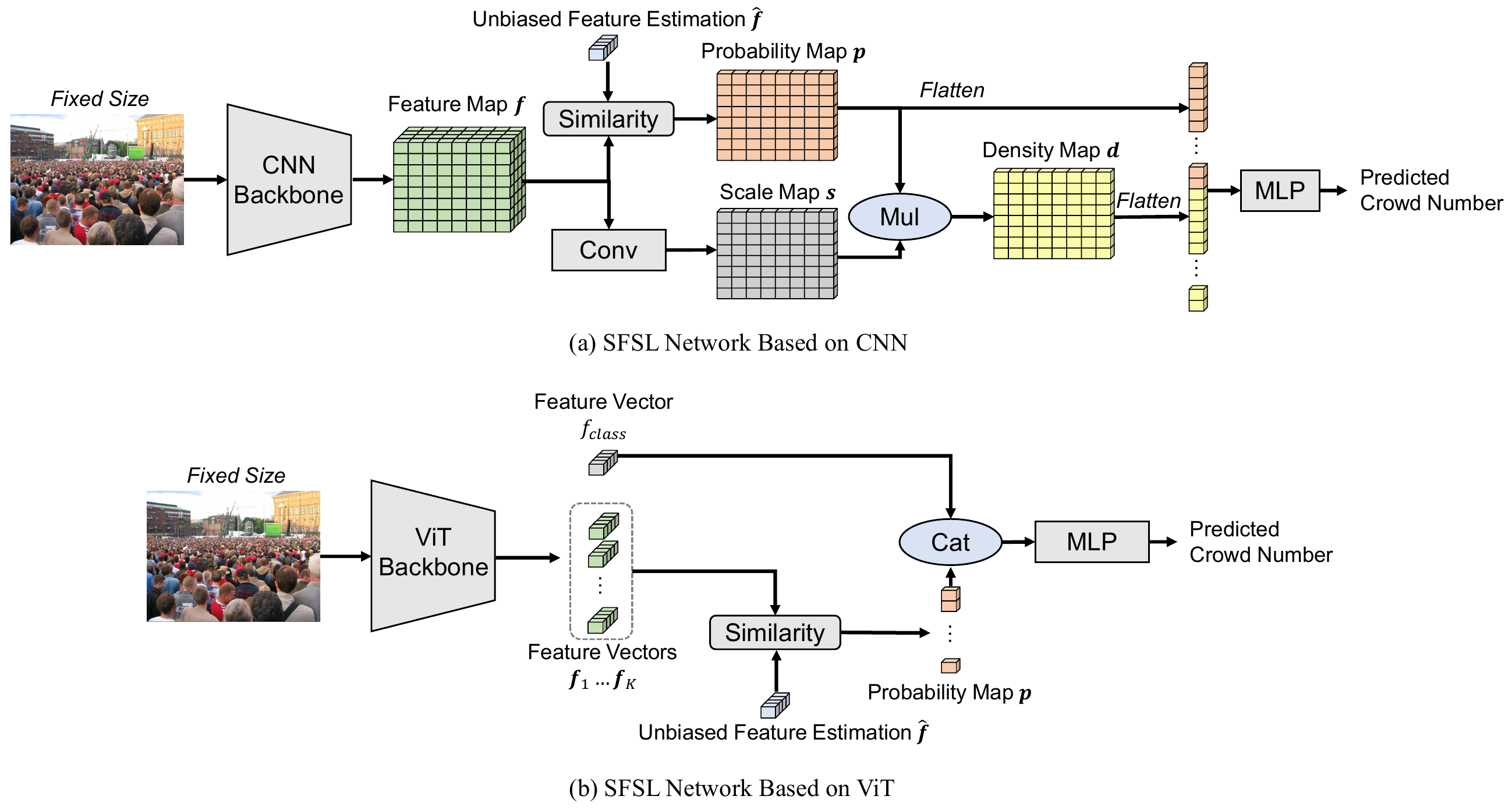}
    \caption{Overview of the self-adaptive feature similarity learning (SFSL) network. (a) is the architecture of the CNN-based SFSL network and (b) is the ViT-based SFSL network. In the CNN-based SFSL network, the density map is generated by multiplying the probability map and the scale map. In both networks, the probability map serves as a part of the MLP input.}
    \label{fig:sfsl}
\end{figure*}

The main challenge of weekly-supervised crowd counting is the lack of local supervision information, because only the total crowd numbers are available.
In order to address this problem, we propose a method for weakly-supervised dense crowd counting to reinforce local feature representation.
First we propose a self-adaptive feature similarity learning (SFSL) network, where a learnable unbiased feature estimation of persons is maintained.
The pixel-wise feature similarity is calculated between the feature vector at each pixel and the unbiased feature estimation vector and thus a similarity map is constructed.
The similarity map is considered as the soft binary classification probability, and is also used as a part of the input of the linear regression network.
Besides, we propose a global-local consistency (GLC) loss to complement the supervision information of the local level.
We establish a constraint on the consistency between the global crowd number and the summation of all the local crowd numbers.

In this section, we elaborate on the details of the two components respectively. The SFSL networks based on CNN and ViT~\cite{dosovitskiy2020image} are introduced separately due to their different architectures.

\subsection{Self-adaptive Feature Similarity Learning}

The prediction of fully-supervised crowd counting is similar with the pixel-wise soft binary classification task.
The estimation at each position can be considered as the probability or the confidence of being classified as the positive sample.
However, there is no local annotation in weakly-supervised crowd counting.
Thus, we propose a self-adaptive similarity learning strategy to reinforce local feature representation.

In the soft binary classification task, there is an unbiased estimation of each class in the feature distribution of the datasets.
Taking the positive class, i.e., the class of persons as an example, the unbiased feature estimation $\hat{\mathbf{f}}$ of this class is formulated as:
\begin{equation}
    \hat{\mathbf{f}} = \mathop{argmax}\limits_{\mathbf{f}}\sum_{i=1}^{M}p_i\mathop{sim}(\mathbf{f}_i, \mathbf{f}),
    \label{unbiased}
\end{equation}
where $M$ is the number of pixels; $\mathbf{f}_i$ and $p_i$ are the feature vector and the probability of being classified as the positive class respectively at the $i$th position; $sim(\cdot)$ is the feature similarity function.

However, it is hard to define $p_i$ in weakly-supervised crowd counting for the lack of local annotations.
In our network, the unbiased feature estimation $\hat{\mathbf{f}}$ of the class of persons is a learnable vector updated self-adaptively by the gradient descent along with the optimization of the network parameters.
Based on the observation of Eq.~\ref{unbiased}, the similarity values tend to be larger at positions with larger $p_i$. 
Given the unbiased feature estimation $\hat{\mathbf{f}}$, we define $p_i$ as
\begin{equation}
    p_i=\lambda_1 \mathop{sim}(\mathbf{f}_i, \hat{\mathbf{f}})
    \label{probability}
\end{equation}
for simplicity, where $\lambda_1$ is a positive hyperparameter.
We adopt the cosine similarity as the feature similarity function and normalize it into range [0, 1], which is calculated as:
\begin{equation}
    \mathop{sim}(\mathbf{f}_i, \hat{\mathbf{f}}) = \frac{\mathbf{f}_i \odot \hat{\mathbf{f}}}{2|\mathbf{f}_i||\hat{\mathbf{f}}|} + 0.5,
    \label{sim}
\end{equation}
where $\odot$ is the inner product of vectors.

\subsubsection{SFSL Network Based on CNN}

The predicted density map is determined not only by the soft binary classification probability, but also by the scales, due to the perspective effect.
As objects with larger receptive fields should be assigned smaller values in the density map, we formulate the density map as:
\begin{equation}
    d_i = \lambda_2\frac{p_i}{s_i} = \lambda_1\lambda_2\frac{\mathop{sim}(\mathbf{f}_i, \hat{\mathbf{f}})}{s_i},
    \label{density}
\end{equation}
where $d_i$ is the density and $s_i$ is the scale at the $i$th position; $\lambda_2$ is another positive hyperparameter.
Both $\lambda_1$ and $\lambda_2$ can be ignored because of the downstream linear regression network.
In practice, the convolutional neural network predicts $\frac{1}{s_i}$ instead of $s_i$ for the output due to the numerical stability, which is formulated as:
\begin{equation}
    \frac{1}{s_i} = F_c(x, \gamma)_i
    \label{density_reverse}
\end{equation}
where $F_c(\cdot)$ denotes the computation of the convolutional layers with the set of parameters $\gamma$.

Both the soft binary classification probability and the density distribution contribute to the final estimation, proved by some fully-supervised works~\cite{shi2019counting,liu2019recurrent,zhao2019leveraging}.
As illustrated in Fig.~\ref{fig:sfsl}, we flatten both the density map and the probability map to vectors for linear regression, denoted as $\mathbf{d}$ and $\mathbf{p}$ respectively, and concatenate them as $(\mathbf{d}, \mathbf{p})$ to construct the input of the linear regression network.

The linear regression network employs three fully-connected layers for the final prediction of the crowd number. This operation is formulated as:
\begin{equation}
    \tilde{c} = F_l((\mathbf{d}, \mathbf{p}), \varphi),
    \label{linear}
\end{equation}
where $\tilde{c}$ is the predicted crowd number, and $F_l(\cdot)$ denotes the linear regression network with the set of parameters $\varphi$.

\subsubsection{SFSL Network based on ViT}

\begin{figure*}[tp]
    \centering
    \includegraphics[width=1.0\linewidth]{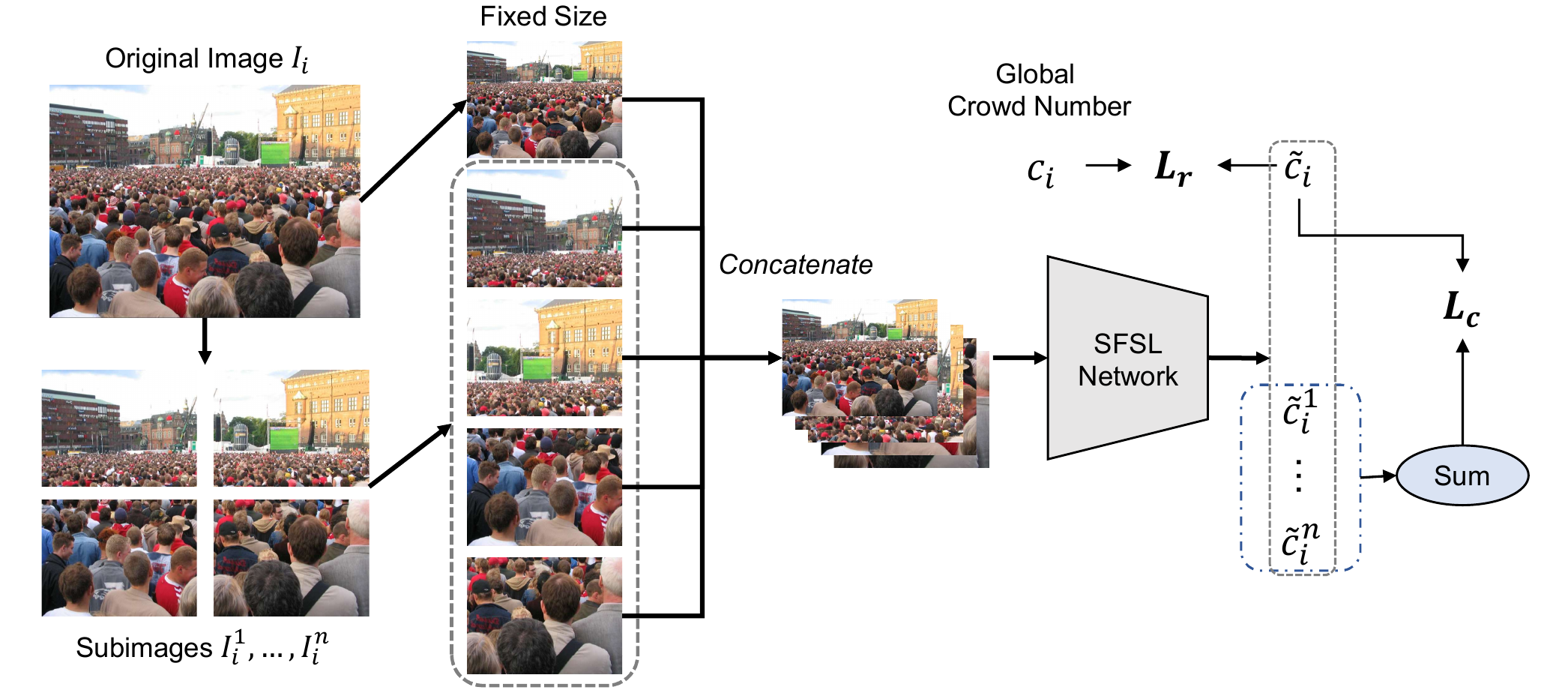}
    \caption{Overview of the global-local consistency loss. The original image is first divided into multiple subimages and then fed into the SFSL network together with the subimages in a batch. The total loss function combines the supervised regression loss $L_r$ and the global-local consistency loss $L_c$.}
    \label{fig:glc}
\end{figure*}

ViT~\cite{dosovitskiy2020image} is the application of Transformer~\cite{vaswani2017attention} in computer vision.
The Transformer encoder~\cite{vaswani2017attention} in ViT~\cite{dosovitskiy2020image} is adopted as the backbone network to substitute the CNN backbone.
Regressing the global crowd numbers by appending a fully-connected layer to ViT~\cite{dosovitskiy2020image} servers as the baseline of our ViT-based SFSL network.

This baseline already integrates the global and local semantic information by the multi-head self-attention layers~\cite{vaswani2017attention} and outputs a vector for downstream predictions.
However, the local feature representation is missing in the MLP decoder, because only the output at the class token serves as the image representation in the decoder.
Unlike other tasks such as image classification, the effectiveness of crowd counting methods is more dependent on the abundance of local semantic information.

As illustrated in Fig.~\ref{fig:sfsl}, the output of the Transformer encoder is a $(K+1)*D$ feature map, where $K$ is the number of image patches and $D$ is the hidden size. $\mathbf{f}_i$ is defined as the feature vector of length $D$ at the $i$th of the $K$ tokens. We calculate the soft binary classification probability $p_i$ following Eq.~\ref{probability}. In the same way as the CNN-based SFSL network, we concatenate the the output vector $\mathbf{d}$ of the original Transformer and the probability vector $\mathbf{p}$ as $(\mathbf{d}, \mathbf{p})$, which serves as the input of the last fully-connected layer. This operation is formulated as:
\begin{equation}
    \tilde{c} = \mathbf{w}^T(\mathbf{d}, \mathbf{p})+b
    \label{vit}
\end{equation}
where $\tilde{c}$ is the predicted crowd number; $\mathbf{w}$ and $b$ are the parameters in the fully-connected output layer.

\subsection{Global-Local Consistency Loss}

In weakly-supervised crowd counting, only the crowd number of the entire image is available.
However, regressing the crowd number at the global level while ignoring the count estimations at the local level may lead to suboptimal predictions.
Conversely, leveraging local regions for predicting the crowd numbers can improve the capability of the network to extract useful and discriminative features.

In the crowd counting task, the global crowd number should be equivalent to the summation of all the local crowd numbers.
Thus there needs to be a constraint on the consistency between them.
We propose a global-local consistency (GLC) loss to enforce this consistency in a self-supervised manner.
As shown in Fig.~\ref{fig:glc}, we divide the original image into multiple subimages, each of which contains a local area without overlapping.
We denote the $i$th image as $I_i$ and the subimages of it as $I_i^1, I_i^2, \dots, I_i^n$, where $n$ is the number of subimages.
We resize these subimages to the same shape as $I_i$ and input the subimages along with $I_i$ into the network simultaneously by concatenating them as $(I_i, I_i^1, I_i^2, \dots, I_i^n)$. In each mini-batch, the input of the network is formulated as $(I_1, I_1^1, I_1^2, \dots, I_1^n, I_2, \dots, I_b^n)$, where $b$ is the batch size.

We adopt the MSE loss as both the supervised regression loss and the GLC loss.
The supervised regression loss is formulated as:
\begin{equation}
    \mathcal{L}_r = \frac{1}{N}\sum_{i=1}^N(\tilde{c}_i - c_i)^2 = \frac{1}{N}\sum_{i=1}^N(F(I_i, \{\gamma,\varphi, \hat{\mathbf{f}}\}) - c_i)^2,
    \label{eq6}
\end{equation}
where $\tilde{c}_i$ is the prediction of the $i$th image $I_i$ from the network $F$ with the parameter set $\{\gamma,\varphi, \hat{\mathbf{f}}\}$.
The GLC loss is formulated as:
\begin{equation}
\begin{aligned}
    &\mathcal{L}_c = \frac{1}{N}\sum_{i=1}^N(\sum_{j=1}^n\tilde{c}_i^j- \tilde{c}_i)^2 \\&= \frac{1}{N}\sum_{i=1}^N(\sum_{j=1}^nF(I_i^j, \{\gamma,\varphi, \hat{\mathbf{f}}\})-F(I_i, \{\gamma,\varphi, \hat{\mathbf{f}}\}))^2,
    \label{eq7}
\end{aligned}
\end{equation}
where $\tilde{c}_i^j$ is the prediction of the $j$th subimage of $I_i$ from the network.
This loss function indicates that the summation of the local predictions is constrained to be close to the global prediction.

The total loss function $\mathcal{L}_{total}$ is comprised of the above two loss terms:
\begin{equation}
    \mathcal{L}_{total} = \mathcal{L}_r + \alpha \mathcal{L}_c,
    \label{eq8}
\end{equation}
where $\alpha$ is the weight of the GLC loss to balance the two loss terms.

An intuitional and superficial strategy to leverage local features is supervising the summation of the local predictions by the ground truth crowd number $c_i$.
This approach is an effective data augmentation because of the concatenation after the cropping and resizing operations.
However, the consistency between the predictions of global and local areas is not maintained.
Another method to leverage the feature representation of local regions is to regress the global crowd numbers first and supervise the local regression by the global predictions afterwards.
The separate operations weaken the supervision of the ground truth on local regions and the inaccurate global predictions impair the local regression.
Distinguished from the above two strategies, our GLC loss not only maintains the accuracy of the global regression the but also makes the network predictions more consistent between scenes of different scales.

\section{Experiments}

\subsection{Datasets}

In this section, we introduce the four datasets which we carry out experiments on.
Note that the location annotations in these datasets are not used in all our experiments, and we label each image with only the total crowd number.

\subsubsection{ShanghaiTech}

\cite{zhang2016single} is divided into two parts named Part A and Part B.
In Part A, there are 482 images from the Internet, including 300 images in the training set and 182 images in the test set.
There are 501 labeled people per image on average.
In Part B, there are 716 images taken from surveillance viewers, including 400 training images and 316 test images.
The crowds in this part are relatively sparse, containing 123 labeled people per image on average.

\subsubsection{UCF-QNRF}

\cite{idrees2018composition} is a large and dense crowd dataset, containing 1201 training images and 334 test images.
There are about 1.25 million annotations in total, an average of 815 labels per image.
The average resolution of this dataset is $2013\times 2902$.

\subsubsection{NWPU-Crowd}

\cite{wang2020nwpu} contains 5,109 images with over 2.13 million annotations.
This dataset includes 3109 images in the training set and 500 images in the validation set.
There is no access to the labels of the 1,500 test images.
The image resolution is extremely large, averaging $2191\times 3209$ per image.

\subsubsection{JHU-CROWD++}

\cite{sindagi2020jhu} is a challenging large-scale dataset that contains 4,372 images with over 1.51 million annotations.
There are 2,272 images for training, 500 images for validation, and 1,600 images for testing.

\begin{table}[tp]
    \centering
    \caption{Ablation studies of the CSRNet\_bn-based network on ShanghaiTech Part A.}
    \begin{threeparttable}
    \begin{tabular}{c|c|c|c}
    \hline
    \noalign{\smallskip}
        Input of MLP\tnote{1} & Loss\tnote{2} & MAE & MSE \\
    \noalign{\smallskip}
    \hline
    \noalign{\smallskip}
        $(\mathbf{d}')$ & $L_r$ & 109.6 & 169.5\\
        $(\mathbf{s}\cdot\mathbf{p}, \mathbf{p})$ & $L_r$ & 94.77 & 139.7\\
        $(\mathbf{s}\cdot\mathbf{p}, \mathbf{p})$ & $L_r + L_{gt} (n=4)$ & 87.17 & 135.2\\
        $(\mathbf{s}\cdot\mathbf{p}, \mathbf{p})$ & $L_r + L_c(n=16)$ & 88.13 & 137.6\\
        $(\mathbf{d}')$ & $L_r + L_c(n=4)$ & 85.55 & 132.6\\
        $(\mathbf{s}\cdot\mathbf{p})$ & $L_r + L_c(n=4)$ & 81.50 & 127.3\\
        $(\mathbf{d}', \mathbf{p})$ & $L_r + L_c(n=4)$ & 82.39 & 127.3\\
        $(\mathbf{s}\cdot\mathbf{p}, \mathbf{p})$ & $L_r + L_c(n=4)$ & \textbf{81.21} & \textbf{127.1}\\
    \hline
    \end{tabular}
    \begin{tablenotes}
    \footnotesize \item [1] $\mathbf{d}'$ denotes that the backbone generates the density map directly without constructing the scale map $\mathbf{s}$ and the probability map $\mathbf{p}$.
    \item [2] $L_{gt}$ is the loss between the summation of local predictions and the ground truth crowd number.
    \end{tablenotes}
    \end{threeparttable}
    \label{tab:ablation_cnn}
\end{table}

\begin{table}[tp]
    \centering
    \caption{Ablation studies of the ViT-based network on ShanghaiTech Part A.}
    \begin{threeparttable}
    \begin{tabular}{c|c|c}
    \hline
    \noalign{\smallskip}
        Method & MAE & MSE\\
    \noalign{\smallskip}
    \hline
    \noalign{\smallskip}
        Baseline\tnote{1} & 84.54 & 127.6\\
        Baseline + SFSL & 83.03 & 126.0\\
        Baseline + GLC loss & 83.07 & 125.9\\
        Baseline + SFSL + GLC loss & \textbf{82.73} & \textbf{122.8}\\
    \hline
    \end{tabular}
    \begin{tablenotes}
    \footnotesize \item [1] The baseline is the original ViT network.
    \end{tablenotes}
    \end{threeparttable}
    \label{tab:ablation_vit}
\end{table}

\begin{table}[tp]
    \centering
    \caption{The effect of our method on different backbones on ShanghaiTech Part A.}
    \begin{threeparttable}
    \begin{tabular}{c|cc|cc}
    \hline
    \noalign{\smallskip}
        \multirow{2}{*}{Backbone\tnote{1}} & \multicolumn{2}{c|}{Baseline} & \multicolumn{2}{c}{Ours} \\
         & MAE & MSE & MAE & MSE\\
    \noalign{\smallskip}
    \hline
    \noalign{\smallskip}
    MCNN\_bn & 165.4 & 248.1 & 149.6 & 234.5\\
    VGG16\_bn & 116.5 & 168.9 & 98.12 & 150.5\\
    CSRNet\_bn & 109.6 & 169.5 & 81.21 & 127.1\\
    ViT & 84.43 & 127.6 & 82.73 & 122.8\\
    \hline
    \end{tabular}
    \begin{tablenotes}
    \footnotesize \item [1] \_bn denotes that we append Batch Normalization layers to the backbone.
    \end{tablenotes}
    \end{threeparttable}
    \label{tab:backbones}
\end{table}

\begin{table*}[tp]
    \centering
    \caption{Comparisons with CNN and ViT baselines on four datasets.}
    \begin{tabular}{c|cc|cc|cc|cc|cc}
    \hline
    \noalign{\smallskip}
        \multirow{2}{*}{Method} & \multicolumn{2}{c|}{STA} & \multicolumn{2}{c|}{STB} & \multicolumn{2}{c|}{QNRF} & \multicolumn{2}{c|}{NWPU val} & \multicolumn{2}{c}{JHU val}\\
        & MAE & MSE & MAE & MSE & MAE & MSE & MAE & MSE & MAE & MSE\\
    \noalign{\smallskip}
    \hline
    \noalign{\smallskip}
    Baseline (CNN) & 109.6 & 169.5 & 19.34 & 33.86 & 187.3 & 310.7 & 151.7 & 531.3 & 130.4 & 380.4\\
    Baseline (ViT) & 84.54 & 127.6 & 16.65 & 27.74 & 153.9 & 262.5 & 135.5 & 555.7 & 96.66 & 266.6\\
    Ours (CNN) & \textbf{81.21} & 127.1 & 16.83 & 28.45 & 161.9 & 273.7 & 138.6 & 512.2 & 126.8 & 376.5\\
    Ours (ViT) & 82.73 & \textbf{122.8} & \textbf{14.93} & \textbf{25.53} & \textbf{145.8} & \textbf{249.0} & \textbf{135.4} & \textbf{512.0} & \textbf{93.50} & \textbf{242.9}\\
    \hline
    \end{tabular}
    \label{tab:ablation}
\end{table*}

\begin{figure*}[tp]
    \centering
    \includegraphics[width=1.0\linewidth]{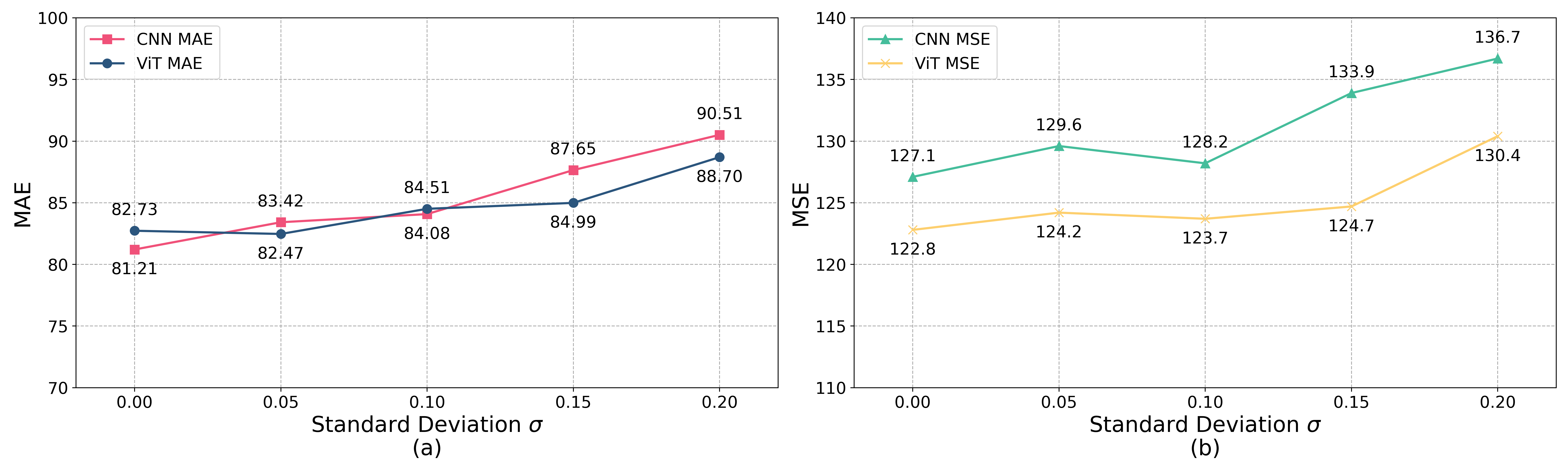}
    \caption{The validation experiments of the robustness against labeling deviations. We test different standard deviations of the Gaussian noise and carry out experiments on ShanghaiTech Part A.}
    \label{fig:robustness}
\end{figure*}

\subsection{Implementation Details}

In weakly-supervised manner, only the global crowd number of an image is available, so sampling data by cropping images is inappropriate for the lack of local ground truth crowd numbers.
In our experiments, we resize all the images to a fixed size.
In the CNN-based network, we resize images to the shape of $256\times 256$, while in the ViT-based network, the shape of resized images is $384\times 384$ to be consistent with ViT~\cite{dosovitskiy2020image}. The random horizontal flipping is adopted as the data augmentation.

In the CNN-based network, the backbone is CSRNet~\cite{li2018csrnet} with Batch Normalization~\cite{ioffe2015batch}, which is labeled as CSRNet\_bn in this paper.
In the ViT-based network, the backbone is ViT-Base~\cite{dosovitskiy2020image}, where there are 12 layers, 12 heads, the hidden size as 768 and the MLP size as 3072.
The first ten layers of the CSRNet\_bn~\cite{li2018csrnet,ioffe2015batch} and the entire ViT~\cite{dosovitskiy2020image} backbone are pretrained on ImageNet~\cite{russakovsky2015imagenet}.

The number of subimages $n$ in the GLC loss is set to $4$, and the we randomly select 6 images from the training set in one batch.
Thus the effective batch size is $(4+1)\times 6=30$.
The weight $\alpha$ of the GLC loss is set to $1$.

We adopt Adam Optimizer~\cite{kingma2014adam} with a fixed learning rate $10^{-5}$ and the weight decay $10^{-4}$ in the training process.

\subsection{Ablation Study}

\begin{table*}[tp]
    \centering
    \caption{Comparisons with state-of-the-art methods on four datasets. The methods are divided into three groups according to the supervision manners, which are fully-supervised, semi-supervised and weakly-supervised manners from top to bottom.}
    \begin{threeparttable}
    \begin{tabular}{c|C{0.2cm}C{0.2cm}C{0.2cm}|C{0.7cm}C{0.7cm}|C{0.7cm}C{0.7cm}|C{0.7cm}C{0.7cm}|C{0.7cm}C{0.7cm}|C{0.7cm}C{0.7cm}}
    \hline
    \noalign{\smallskip}
        \multirow{2}{*}{Method} & \multicolumn{3}{c|}{Labels\tnote{1}} & \multicolumn{2}{c|}{STA} & \multicolumn{2}{c|}{STB} & \multicolumn{2}{c|}{QNRF} & \multicolumn{2}{c|}{NWPU test} & \multicolumn{2}{c}{JHU test}\\
         & I & P & L & MAE & MSE & MAE & MSE & MAE & MSE & MAE & MSE & MAE & MSE\\
    \noalign{\smallskip}
    \hline
    \noalign{\smallskip}
    MCNN(2016)~\cite{zhang2016single} & \Checkmark & \Checkmark & \Checkmark & 110.2 & 173.2 & 26.4 & 41.3 & 277 & 426 & 232.5 & 714.6 & 188.9 & 483.4\\
    CMTL(2017)~\cite{sindagi2017cnn} & \Checkmark & \Checkmark & \Checkmark & 101.3 & 152.4 & 20.0 & 31.1 & 252 & 514 & - & - & 157.8 & 490.4\\
    CSRNet(2018)~\cite{li2018csrnet} & \Checkmark & \Checkmark & \Checkmark & 68.2 & 115.0 & 10.6 & 16.0 & - & - & 121.3 & 387.8 & 85.9 & 309.2\\
    TEDNet(2019)~\cite{jiang2019crowd} & \Checkmark & \Checkmark & \Checkmark & 64.2 & 109.1 & 8.2 & 12.8 & 113 & 188 & - & - & - & -\\
    ADSCNet(2020)~\cite{bai2020adaptive} & \Checkmark & \Checkmark & \Checkmark & 55.4 & 97.7 & 6.4 & 11.3 & \textbf{71.3} & \textbf{132.5} & - & - & - & - \\
    Wan et al.(2021)~\cite{wan2021generalized}& \Checkmark & \Checkmark & \Checkmark & 61.3 & 95.4 & 7.3 & 11.7 & 84.3 & 147.5 & 79.3 & \textbf{346.1} & \textbf{59.9} & \textbf{259.5}\\
    P2PNet(2021)~\cite{song2021rethinking} & \Checkmark & \Checkmark & \Checkmark & \textbf{52.74} & \textbf{85.06} & \textbf{6.25} & \textbf{9.9} & 85.32 & 154.5 & \textbf{77.44} & 362 & - & -\\
    \hline
    IRAST(2020)~\cite{liu2020semi} & \Checkmark & \Checkmark & \Checkmark & 86.9 & 148.9 & 14.7 & 22.9 & 135.6 & 233.4 & - & - & - & -\\
    SUA(2021)~\cite{meng2021spatial} & \Checkmark & \Checkmark & \Checkmark & \textbf{68.5} & 121.9 & 14.1 & 20.6 & 130.3 & 226.3 & \textbf{111.7} & \textbf{443.2} & \textbf{80.7} & \textbf{290.8}\\
    Xu et al.(2021)~\cite{xu2021crowd} & \XSolidBrush & \Checkmark & \Checkmark & 72.79 & \textbf{111.6} & \textbf{12.03} & \textbf{18.70} & \textbf{128.1} & \textbf{218.1} & 178.7 & 1080 & 129.7 & 400.5\\
    \hline
    Yang et al.(2020)~\cite{yang2020weakly} & \Checkmark & - & \XSolidBrush & 104.6 & 145.2 & 12.3 & 21.2 & - & - & - & - & - & -\\
    MATT(2021)~\cite{lei2021towards} & \Checkmark & \Checkmark & \Checkmark & 80.1 & 129.4 & 11.7 & 17.5 & - & - & - & - & - & -\\
    TransCrowd(2021)~\cite{liang2021transcrowd} & \Checkmark & \Checkmark & \XSolidBrush & \textbf{66.1} & \textbf{105.1} & \textbf{9.3} & \textbf{16.1} & \textbf{97.2} & \textbf{168.5} & \textbf{117.7} & 451.0 & \textbf{74.9} & \textbf{295.6}\\
    Ours (CNN) & \Checkmark & \XSolidBrush & \XSolidBrush & 81.21 & 127.1 & 16.83 & 28.45 & 161.9 & 273.7 & 159.1 & 508.4 & 143.9 & 485.3\\
    Ours (ViT) & \Checkmark & \XSolidBrush & \XSolidBrush & 82.73 & 122.8 & 14.93 & 25.53 & 145.8 & 249.0 & 137.4 & \textbf{425.6} & 116.5 & 404.4\\
    \hline
    \end{tabular}
    \begin{tablenotes}
    \footnotesize \item [1] ``I" denotes the image-wise crowd numbers; ``P" denotes the patch-wise crowd numbers; ``L" denotes the locations of objects.
    The mark ``-" indicates that whether the labels are available is unknown.
    \end{tablenotes}
    \end{threeparttable}
    \label{tab:sota}
\end{table*}

\begin{table*}[tp]
    \centering
    \caption{The results of methods with different amounts of labels. Our model trained with patch-wise labels is denoted as ``Ours*".}
    \begin{tabular}{c|c|cc|cc}
    \hline
    \noalign{\smallskip}
        \multirow{2}{*}{Method} & \multirow{2}{*}{Labels} & \multicolumn{2}{c|}{STA} & \multicolumn{2}{c}{STB} \\
        & & MAE & MSE & MAE & MSE\\
    \noalign{\smallskip}
    \hline
    \noalign{\smallskip}
    TransCrowd~\cite{liang2021transcrowd} & $6\times$ & \textbf{66.1} & 105.1 & 9.3 & 16.1\\
    Ours (ViT) & $1\times$ & 82.73 & 122.8 & 14.93 & 25.53\\
    Ours* (ViT) & $6\times$ & 68.81 & \textbf{103.9} & \textbf{9.248} & \textbf{14.22}\\
    \hline
    \end{tabular}
    \label{tab:6patch}
\end{table*}

In this subsection, we perform some ablation studies to analyze the proposed SFSL network and the GLC loss, including the validation of the effectiveness on different backbones.

The ablation studies of the network based on CSRNet\_bn~\cite{li2018csrnet,ioffe2015batch} on the ShanghaiTech~\cite{zhang2016single} Part A dataset are shown in Tab.~\ref{tab:ablation_cnn}.
First we test three configurations of the input of the MLP layer.
The baseline generates the density map $\mathbf{d}'$ directly, while in our SFSL network, the scale map $\mathbf{s}$ is the output of convolutional layers and the probability map $\mathbf{p}$ is generated by SFSL.
It is observed that both generating the density map by SFSL and combining the probability map in the input of MLP can improve the performance.
Then we compare the performance of different loss functions, including the regression loss $L_r$, our GLC loss $L_c$ and the loss between the summation of local predictions and the global ground truth.
The proposed GLC loss outperforms the other loss functions.
We also enlarge the number of subimages $n$ in the GLC loss from $2\times 2=4$ to $4\times 4=16$.
However, excessive subimages make it more difficult to maintain the consistency between the global and the local feature representations.
There is more noise in local predictions for the absence of local labels when there are more subimages.
To balance the trade-off, we set the number of subimages $n$ to $4$ in our experiments.

We also validate the performance of our method based on the ViT~\cite{dosovitskiy2020image} backbone, and the results are shown in Tab.~\ref{tab:ablation_vit}.
It is demonstrated that both our SFSL network and GLC loss improve the performance in terms of both the MAE and MSE metrics, and the two components of our method is compatible.

To verify the expansion capability, we introduce our method to boost the performance of other backbones such as MCNN~\cite{zhang2016single} and VGG16~\cite{simonyan2014very}.
To exclude the interference from irrelevant factors, we use the same experiment setting including Batch Normalization~\cite{ioffe2015batch} for all the backbones.
As shown in Tab.~\ref{tab:backbones}, our method boosts the performance of all the four backbones with consistent improvements.
Our method decreases the MAE metric by 9.6\%, 15.8\% and 25.9\% respectively on the three CNN backbones, which verifies the effectiveness of our proposed method.
This percentage is 2.0\% on the ViT~\cite{dosovitskiy2020image} backbone.
The relatively small improvement can be attributed to the more powerful capability of leveraging local semantic information and fusing local feature representations in ViT~\cite{dosovitskiy2020image}, compared with CNN.
Nevertheless, our method can further boost ViT~\cite{dosovitskiy2020image} by reinforcing local feature representation, which proves that our method is compatible with other methods related to the correlation among local areas.

We evaluate our method on four datasets, including ShanghaiTech Part A and Part B, UCF-QNRF and NWPU.
The comparison experiments on these datasets are reported in Tab.~\ref{tab:ablation}, where the CNN-based network adopts CSRNet\_bn~\cite{li2018csrnet,ioffe2015batch} as the backbone.
These experiments demonstrate that our method is applicable for different crowd scenes considering the crowd density, the perspective effect and the object scales.

\subsection{Robustness Evaluation}

In weakly-supervised crowd counting, the ground truth crowd number of an image can be obtained by economical methods such as crowd sensing technology~\cite{guo2015mobile}, instead of manual annotations.
These methods can reduce the workload of annotations significantly, but introduce labeling deviations due to inaccurate annotations.
Thus it is necessary to validate the robustness of our method when encountering different levels of labeling deviations.

We adopt the Gaussian noise as the simulated labeling deviations in our experiments.
The ground truth crowd number of an image with simulated noise is formulated as:
\begin{equation}
    c' = c(1+\epsilon), \epsilon \sim N(0, \sigma^2),
    \label{eq:noise}
\end{equation}
where $c$ is the original ground truth crowd number; $N(0, \sigma^2)$ is the Gaussian distribution with the mean of $0$ and the variance of $\sigma^2$.
We use different $\sigma$ from $0.05$ to $0.2$ in our experiments.

The results of the robustness evaluation experiments are illustrated in Fig.~\ref{fig:robustness}.
It is observed that there is no obvious loss of prediction accuracy with simulated noise in the ground truth crowd number, especially when the standard deviation of the Gaussian noise is smaller than $0.1$.
The MAE only increases by 3.5\% and 2.2\% and the MSE only increases by 0.9\% and 0.7\% with the CNN-based and the ViT-based networks respectively, when $\sigma=0.1$.
The inspiring results prove the tolerance of labeling deviations in our weakly-supervised method and provide strong support for the reduction of manual annotations.

\subsection{Comparisons with State-of-the-Arts}

We compare our proposed method with state-of-the-art methods, including fully-supervised, semi-supervised and weakly-supervised methods.
The results of our comparison experiments on four datasets are shown in Tab.~\ref{tab:sota}.
The compared methods are divided into three categories according to the supervision manners.
Besides, we also divide the training labels into three levels: image level, patch level and pixel level.
The image-level labels are the image-wise crowd numbers, i.e., the global ground truth in our method, and the pixel-level labels are the locations of objects.
In some weakly-supervised methods, cropped images serve as the input of the network and the crowd number in each image patch is available in the training process.
Thus it is defined that patch-level labels are used in these methods.

In our data augmentation, there is no image cropping process.
The only available labels are the image-wise crowd numbers.
For example, in the ShanghaiTech Part A dataset~\cite{zhang2016single}, only 300 crowd numbers serve as the training labels, for there are 300 images in the training set.
Obviously, although our method achieves competitive results, the insufficient quantity of annotations limits the performance of our method.
To evaluate the performance of our method under the circumstance of more annotations, we also test our method with patch-wise labels.
Following TransCrowd~\cite{liang2021transcrowd}, we crop each image in training sets into 6 patches and label each patch with the patch-wise crowd number.
In this way, the quantity of annotations increases to 6 times.
The results on the ShanghaiTech dataset~\cite{zhang2016single} are shown in Tab.~\ref{tab:6patch}.
Our model trained with patch-wise labels is denoted as ``Ours*".

The experiments in Tab.~\ref{tab:6patch} demonstrate that there is much room for performance improvement in our method when there are more images labeled with the total crowd numbers in the training set.
In fact, without the locations of objects in the weakly-supervised manner, the annotation work is much less laborious.
Therefore, enlarging the quantity of the labels is effortless and our method has the potential to achieve better performance with more labeled data.

\section{Conclusion}

We propose a self-adaptive feature similarity learning (SFSL) network and a global-local consistency (GLC) loss to reinforce local feature representation in weakly-supervised crowd counting.
Our method reduces the workload of annotations significantly and maintains good performance on four crowd counting datasets.
We also demonstrate the tolerance of labeling deviations in our weakly-supervised method, which provides strong support for the reduction of manual annotations.
Besides, with more images labeled with the total crowd numbers, our method can achieve much higher counting accuracy.

In terms of the effectiveness, robustness to labeling deviations and the potential to boost performance with more crowd numbers, our method narrows the gap between weakly-supervised and fully-supervised dense crowd counting.

{\small
\bibliographystyle{ieee_fullname}
\bibliography{paper}
}

\end{document}